\documentclass[manuscript]{acmart}

\AtBeginDocument{%
  \providecommand\BibTeX{{%
    \normalfont B\kern-0.5em{\scshape i\kern-0.25em b}\kern-0.8em\TeX}}}

\setcopyright{acmcopyright}
\copyrightyear{2018}
\acmYear{2018}
\acmDOI{XXXXXXXXXXXXXX.XXXXXXXXXXXXXX}

\acmPrice{15.00}
\acmISBN{978-1-4503-XXXX-X/18/06}

\usepackage{amsmath}
\usepackage{array}

\usepackage{graphicx}
\usepackage{algorithm}
\usepackage{algorithmic}
\usepackage{booktabs}
\usepackage{multirow}
\usepackage{amsfonts}

\usepackage{nicefrac}
\usepackage{wrapfig}
\usepackage{hyperref}

\makeatletter
\newcommand{\ssymbol}[1]{$^{\@fnsymbol{#1}}$}
\makeatother

\newcommand{\R}{\mathbb{R}}

\begin{document}

\title{Rethinking and Improving Natural Language Generation with Layer-Wise Multi-View Decoding}

\author{Fenglin Liu}
\authornote{Both authors contributed equally to this research.}
\affiliation{
  \institution{Department of Engineering Science, University of Oxford}
  \country{UK}
}
\email{fenglinliu98@gmail.com}

\author{Xuancheng Ren}
\authornotemark[1]
\affiliation{
  \institution{Department of Computer Science, Peking University}
  \country{China}
}
\email{renxc@pku.edu.cn}

\author{Guangxiang Zhao}
\affiliation{
  \institution{Department of Computer Science, Peking University}
  \country{China}
}
\email{zhaoguangxiang@pku.edu.cn}

\author{Chenyu You}
\affiliation{
  \institution{Department of Electrical Engineering, Yale University}
  \country{USA}
}
\email{chenyu.you@yale.edu}

\author{Xuewei Ma}
\affiliation{
  \institution{School of ECE, Peking University}
  \country{China}
}
\email{maxuewei971201@stu.pku.edu.cn}

\author{Xian Wu}
\affiliation{
  \institution{Tencent AI Lab}
  \country{China}
}
\email{kevinxwu@tencent.com}

\author{Xu Sun}
\affiliation{
  \institution{Department of Computer Science, Peking University}
  \country{China}
}
\email{xusun@pku.edu.cn}

\renewcommand{\shortauthors}{Liu and Ren, et al.}

\begin{abstract}
In sequence-to-sequence learning, e.g., natural language generation, the decoder relies on the attention mechanism to efficiently extract information from the encoder. While it is common practice to draw information from only the last encoder layer, recent work has proposed to use representations from different encoder layers for diversified levels of information. Nonetheless, the decoder still obtains only a single view of the source sequences, which might lead to insufficient training of the encoder layer stack due to the \textit{hierarchy bypassing} problem. In this work, we propose \textit{layer-wise multi-view decoding}, where for each decoder layer, together with the representations from the last encoder layer, which serve as a global view, those from other encoder layers are supplemented for a stereoscopic view of the source sequences. Systematic experiments and analyses show that we successfully address the hierarchy bypassing problem, require almost negligible parameter increase, and substantially improve the performance of sequence-to-sequence learning with deep representations on five diverse tasks, i.e., machine translation, abstractive summarization, image captioning, video captioning, medical report generation, and paraphrase generation. In particular, our approach achieves new state-of-the-art results on ten benchmark datasets, including a low-resource machine translation dataset and two low-resource medical report generation datasets.\footnote{The codes have been attached to supplementary material and will be released publicly upon publication.}
\end{abstract}




\maketitle

\section{Introduction}

In recent years, encoder-decoder based models \cite{Bahdanau2015seq2seq,Vaswani2017transformer,Xu2015Show} have become the fundamental instrument for sequence-to-sequence learning, especially in tasks that involve natural language. 
The attention mechanism \cite{Luong2015Effective,Bahdanau2015seq2seq} proves essential for the encoder-decoder based models to efficiently draw useful source information from the encoder.

\begin{figure*}
	\begin{minipage}{0.43\linewidth}
		 \centering
	    \includegraphics[width=0.7\linewidth]{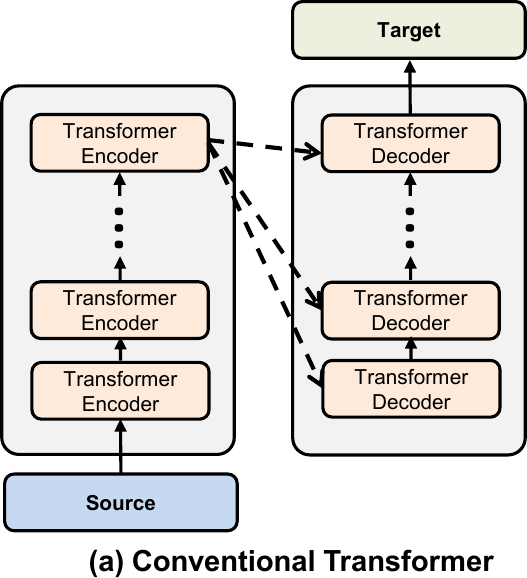}
		\makeatletter\def\@captype{figure}\makeatother
        \vspace{-10pt}
		\caption{Illustration of the conventional Transformer \cite{Vaswani2017transformer}.
		\label{fig:compare}}
	\end{minipage}
	\hfill
	\begin{minipage}{0.52\linewidth}
	    \centering
	    \footnotesize
		\makeatletter\def\@captype{table}\makeatother
		\caption{Examples of the target sentence translated by different methods. Base stands for the conventional Transformer-Base \cite{Vaswani2017transformer} which generates unfaithful and repeated text.
        \label{tab:examples}}
        \vspace{5pt}
        \begin{tabular}{|m{0.1\linewidth}|m{0.68\linewidth}|}
        \hline
        \ \textbf{Source}    &  aber ich hatte das gro\ss{}e gl\"uck , ihn sehr fr\"uh , mit \textit{dreizehn} jahren , kennenzulernen und so war ich schon zu meiner schulzeit auf seinen kursen .  \\ \hline
        
        \ \textbf{Target}    &but i had the good fortune to meet him at a very young age , when i was \textit{thirteen} , and so i always  attended his courses while i was at school .             \\ \hline
        
        \ \textbf{Base} & but i was very lucky to meet him\textit{ at the age of three at the age of three} , and so at the time , i was  on his classes .           \\ \hline
        
        \ \textbf{Ours} &  but i was very lucky to meet him at the age of \textit{thirteen}, and so at the time while i was at school , i always on his courses .      \\ \hline
        \end{tabular}
	\end{minipage}
\end{figure*}

As illustrated in Figure~\ref{fig:compare}, for those representations output by different encoder layers, the common practice for the decoder is to draw information only from the last encoder layer, which is regarded as a global and comprehensive view of the source sequence but short of precise and finer details. The impact of such architectural choice is demonstrated in Table~\ref{tab:examples}. For this German-to-English translation example, the target sentence generated by the system reveals two common shortcomings in neural sequence-to-sequence models: 1) the generated text is unfaithful to the source (e.g., \textit{three} compared to \textit{dreizehn ``thirteen''}); and 2) repeated texts are generated (e.g., \textit{at the age of three}) \cite{see2017Point}. They can both be attributed to the lack of detailed and accurate information.

Several studies have tried to narrow the information gap by introducing deep representations from the encoder layers. For example, fusing the representations from different encoder layers \cite{dou2018exploiting,Wang2018MLRF,Wang2019Learning} or providing representations from different encoder layers to different decoder layers \cite{Bapna2018Training}. However, mixed conclusions have been reached by such work, either with slight improvements \cite{Bapna2018Training} or obvious degradation \cite{Domhan2018How,He2018Layer-Wise}.
In this work, we identify the \textit{hierarchy bypassing} problem that affects all previous related efforts, hindering the efficient training of the encoder layer stack  \cite{Jeffrey2020Solving} and weakening the hierarchy and diversity of representations from different encoder layers, which is the premise for sequence-to-sequence learning with deep representations.

We propose a novel approach called \textit{layer-wise multi-view decoding}, where for each decoder layer, together with the global view from the last encoder layer, another purposeful view of the source sequence is also supplemented (\textit{soft integration}). Moreover, the layer-wise multi-view training is devised to continue the conventional training so that the representational ability of the multi-layer encoder is directly inherited (\textit{continued learning}). The effect of our approach is systematically investigated on typical strategies of routing source representations (see Figure~\ref{fig:model}), which shows that the proposed approach successfully addresses the hierarchy bypassing problem, requires minimal parameter increase, and improves the performance of sequence-to-sequence learning with deep representations.

Specifically, through our experiments on the machine translation with two strong baselines, i.e., Transformer \cite{Vaswani2017transformer} and DynamicConv \cite{Dou2019Dynamic}, which are the previous state-of-the-art models, we find that one of the multi-view variant, i.e., granularity consistent attention (GCA) (see Figure~\ref{fig:model} (a)), promotes the performance substantially. We speculate the reason is that the GCA builds connections between the corresponding layers in the encoder and the decoder, so that the first decoder layer pays attention to the global information, i.e., \textit{coarse-grained} representations, of the source sequence, which is instrumental in language modeling, while the last decoder layer pays attention to the \textit{fine-grained} representations of the source sequence, which is helpful to generate words that are more precise. Notably, a similar connection pattern without the global view is used in computer vision for biomedical image segmentation with success \cite{U-Net}. It is also considered by \cite{Domhan2018How} in natural language processing but with opposite results, i.e., performance degradation. The reason is that a \textit{direct transfer} of such pattern is not viable for sequence-to-sequence learning and further considerations are required for successful training as we show.

Overall, our contributions are as follows:
\begin{itemize}
    \item \smallskip We propose layer-wise multi-view decoding to efficiently use the diverse-grained representations from the multi-layer encoder. For a decoder layer, together with the global view from the last encoder layer, another purposeful view is supplemented for a stereoscopic view of the source sequences.
    
    \item \smallskip The global view of the source sequence in multi-view decoding mitigates the hierarchy bypassing problem and the proposed continued learning transfers the representational ability of the conventional training seamlessly into multi-view training.
    
    \item \smallskip Extensive experiments and analyses demonstrate that our approach works for representative models and tasks in natural language generation, i.e., Transformer and DynamicConv on the neural machine translation task, consistently and substantially improves the performance of sequence-to-sequence learning with deep representations.
    
    \item \smallskip  We further prove the effectiveness of our approach on abstractive summarization, image captioning, video captioning, medical report generation tasks, and paraphrase generation. The results on ten benchmark datasets show that our approach reaches new state-of-the-art results with almost negligible parameter increase.
\end{itemize}

The rest of the paper is organized as follows: Section~\ref{sec:related} reviews the related works; Section~\ref{sec:approach} introduces the proposed approach; Section~\ref{sec:experiment} and Section~\ref{sec:analysis} present the experimental results and analysis, respectively; Finally, Section~\ref{sec:conclusion} concludes the paper.

\section{Related Work}
\label{sec:related}
In this section, we will describe the related works from 1) Sequence-to-Sequence Learning, 2) Using Deep Representations, 3) Text-based Natural Language Generation, and 4) Vision-based Natural Language Generation.

\subsection{Sequence-to-Sequence Learning} In recent years, many deep neural systems have been proposed for sequence-to-sequence learning. The commonly-used approaches \cite{Bahdanau2015seq2seq,Vaswani2017transformer} depend on the encoder-decoder framework to map a source sequence to a target sequence, such as in machine translation and summarization.
The encoder network computes intermediate representations for the source sequence and the decoder network defines a probability distribution over target sentences given that intermediate representation.
Specifically, to allow a more efficient use of the source sequences, a series of attention methods \cite{Vaswani2017transformer,Xu2015Show,Luong2015Effective} have been proposed to directly provide the decoder with source information. Especially, the recent advent of fully-attentive models, e.g., Transformer \cite{Vaswani2017transformer}, in which no recurrence is required, has been proposed and successfully applied to multiple tasks, e.g., neural machine translation.
The work on attention reveals that attention is efficient, necessary, and powerful at combining information from diverse sources. 
Despite their dominance in the last few years, little work has been done discussing the effect of the connection between the encoder and the decoder in sequence-to-sequence learning.

\subsection{Using Deep Representations} 
In natural language processing, several efforts \cite{Peters2018ELMo,Shen2018Dense,Wang2018MLRF,dou2018exploiting,Domhan2018How,Bapna2018Training,Dou2019Dynamic,Wang2019Learning,Li2020Neuron} have investigated strategies to make the best use of deep representations among layers, e.g., using linear combination \cite{Peters2018ELMo}, dense connection \cite{Shen2018Dense} and hierarchical layer aggregation \cite{dou2018exploiting}. However, most of them \cite{Wang2018MLRF,dou2018exploiting,Dou2019Dynamic,Wang2019Learning,Li2020Neuron} focused on the information within the encoder or the decoder and excluded the effect of the information flow from the encoder to the decoder, which is the focus of this work. Some of them \cite{He2018Layer-Wise,Domhan2018How,Bapna2018Training} considered individual source-to-target attention routing strategy in non-comparable contexts and we provide a unified, systematic overview of such efforts and frame them under the problem of narrowing decoder-encoder information gap. More importantly, they still adopted the single-view decoding approach and faced the hierarchy bypassing problem analyzed in this work, leading to unpromising results in common settings. The multi-view decoding approach is unique to previous work and substantially improve the performance of sequence-to-sequence learning using deep representations.
In brief, in our work:
\begin{itemize}
    \item\smallskip (1) We provide a unified, systematic overview of such efforts and frame them under the problem of narrowing the decoder-encoder information gap;
    
    \item\smallskip (2) We identify the hierarchy bypassing problem that affects all previous related efforts, leading to unpromising results in common settings;
    
    \item\smallskip (3) We further propose the multi-view decoding approach. The extensive experiments show that our approach can improve the performances of different tasks across multiple datasets.
\end{itemize}

\subsection{Vision-based Natural Language Generation}
The task of vision-based natural language generation, including image-based visual captioning, video-based visual captioning, and medical image-based visual captioning (a.k.a. medical report generation), aims to generate a textual description for an input vision content, thus this task combines vision content understanding and language generation and is a cross-modal setting compared to text-based natural language generation. The vision-based natural language generation task belongs to the sequence-to-sequence task, where the source sequence and target sequence contain non-ordered visual features and generated captions, respectively.

Image captioning aims to understand the given images and generate corresponding descriptive sentences \cite{chen2015microsoft}.
The task combines image understanding and language generation.
In recent years, a large number of encoder-decoder based neural systems have been proposed for image captioning \cite{Cornia2020M2,Pan2020XLinear,Pei2019Memory,Venugopalan2015VC,Vinyals2015Show,Xu2015Show,rennie2017self,lu2017knowing,anderson2018bottom,liu2019GLIED,liu2020prophet}.
Specifically, the encoder network \cite{Krizhevsky2012CNN,he2016deep} computes visual representations for the visual contents and the decoder network \cite{Hochreiter1997LSTM,Vaswani2017transformer} generates a target sentence based on the visual representations.
However, the sentence generated by image captioning is usually short and describes the most prominent visual contents, which cannot fully represent the rich feature information of the image.
Beyond the traditional image captioning task, image paragraph generation that produces a long and semantic-coherent paragraph to describe the input image has recently attracted increasing research interests \cite{Krause2017Hierarchical,Liang2017Hierarchical,Yu2016Hierarchical}. To this end, a hierarchical recurrent network (HRNN) \cite{Krause2017Hierarchical,Liang2017Hierarchical} is proposed. In particular, the HRNN uses a two-level RNN model to generate the paragraph based on the image features extracted by a CNN. The two-level RNN includes a paragraph RNN and a sentence RNN, where the paragraph RNN is used to generate topic vectors and each topic vector is used by the sentence RNN to produce a sentence to describe the image. However, the correctness of generating abnormalities should be emphasized more than other normalities in a radiology report, while in a natural image paragraph each sentence has equal importance.

The task of video captioning, which aims to generate a descriptive sentence based on the input video, has a wide range of applications.
Recently, a large number of encoder-decoder based neural models have been proposed for video captioning \cite{Venugopalan2015vc3,Yao2015vc4,Pan2016vc1,Xu2017MA-LSTM,Aafaq2019GRU-EVE,Zheng2020SAAT}.
These methods mainly introduce a convolutional neural network (CNN) \cite{Krizhevsky2012CNN} to encode the video and employ a LSTM \cite{Hochreiter1997LSTM} or a Transformer \cite{Zhou2018Transformer} to generate the coherent captions with the attention mechanism \cite{Bahdanau2015seq2seq,Pan2016vc1}.

Inspired by the success of deep learning models on image captioning, a lot of encoder-decoder based frameworks have been proposed \cite{Jing2018Automatic,Jing2019Show,liu2021PPKED,Liu2019Clinically,Yuan2019Enrichment,Xue2018Multimodal,Li2018Hybrid,Li2019Knowledge,Zhang2020When,Chen2020Generating}.
Specifically, \citet{Jing2018Automatic} proposed a hierarchical LSTM with the attention mechanism \cite{Bahdanau2015seq2seq,Xu2015Show}. \citet{Yuan2019Enrichment} further incorporated the medical concept to enrich the decoder with descriptive semantics. \citet{Xue2018Multimodal} proposed a multimodal recurrent model containing an iterative decoder with visual attention to improve the coherence between sentences.
\citet{Miura2021Factual} proposed an Exact Entity Match Reward and an Entailing Entity Match Reward to improve the factual completeness and consistency of the generated reports, resulting in significant improvements on clinical accuracy.
\citet{Jing2019Show,Li2018Hybrid,Liu2019Clinically} and \citet{Zhang2020When,Li2019Knowledge,liu2021PPKED} introduced the reinforcement learning and medical knowledge graph for chest X-ray report generation, respectively.
However, some errors occur in the generated reports of the existing methods, like duplicate reports, inexact descriptions, etc \cite{Xue2018Multimodal,Yuan2019Enrichment}.

\section{Approach}
\label{sec:approach}

In this section, we will first briefly review the conventional encoder-decoder model \cite{Vaswani2017transformer} and then introduce the proposed multi-view decoding realized on the Transformer model \cite{Vaswani2017transformer}.

\subsection{Conventional Single-View Decoding}
In the conventional encoder-decoder model, the encoder encodes the source sequence $\text{S}_0$ with $N$ stacked encoder layers, defined as 
\begin{equation} 
\text{S}_{i} =f_{\text {encoder}}\left(\text{S}_{i-1}\right).
\end{equation}
By repeating the same process for $N$ times, the encoder outputs the representations $\textbf{S}=\{\text{{S}}_1, \text{{S}}_2, \ldots, \text{{S}}_{N}\}$ of the source sequence. Transformer \cite{Vaswani2017transformer} proposes Multi-Head Attention (MHA) and Feed-Forward Network (FFN) to implement the encoder layers:
\begin{equation} 
f_{\text {encoder}}\left(\text{x}\right) = \text{FFN}\left(\text{MHA}\left(\text{x,x,x}\right)\right).
\end{equation}
The MHA consists of $n$ parallel heads and each head is defined as a scaled dot-product attention:
\begin{align}
&\text{Att}_i(\text{x,y,y}) = \text{softmax}\left(\frac{\text{x}\text{W}_i^\text{Q}(\text{y}\text{W}_i^\text{K})^T}{\sqrt{{d}_{n}}}\right)\text{y}\text{W}_i^\text{V},
& \text{MHA}(\text{x,y,y}) = [\text{Att}_1(\text{x,y,y}); \dots; \text{Att}_n(\text{x,y,y})]\text{W}^\text{O}
\end{align}
where $\text{x} \in \R^{l_x \times d}$ and $\text{y} \in \R^{l_y \times d}$ denote the Query matrix and the Key/Value matrix, respectively; $\text{W}_i^\text{Q}, \text{W}_i^\text{K}, \text{W}_i^\text{V} \in \R^{d \times d_n}$ and $\text{W}^\text{O} \in \R^{d \times d}$ are learnable parameters, where ${d}_{n} = d / {n}$; $[\cdot;\cdot]$ stands for concatenation operation.

Following the MHA is the FFN, defined as follows:
\begin{align}
\text{FFN}(\text{x}) = \max(0,\text{x}\text{W}_\text{f}+\text{b}_\text{f})\text{W}_\text{ff}+\text{b}_\text{ff} 
\end{align}
where $\max(0,*)$ represents the ReLU activation function; $\text{W}_\text{f} \in \R^{d \times 4d}$ and $\text{W}_\text{ff} \in \R^{4d \times d}$ denote  learnable matrices for linear transformation; $\text{b}_\text{f}$ and $\text{b}_\text{ff}$ represent the bias terms. It is worth noting that both the MHA and FFN are followed by an operation sequence of dropout \cite{dropout}, residual connection \cite{he2016deep}, and layer normalization \cite{Ba2016layernorm}.

Then, the conventional decoder generates the target sequence $\text{T}_N$ with $N$ stacked decoder layers. 
Typically, only the representations from the final encoder layer  $\text{S}_N$ serve as the bridge between the source and the target, via the attention mechanism.
Unlike the encoder, the decoder has an extra objective that handles the intermediate representations from the encoder, defined as 
\begin{equation} 
\text{T}_{i} =f_{\text {decoder}}\left(\text{T}_{i-1}, \text{S}_{N}\right). \label{eq:dec} 
\end{equation}
For Transformer, the decoder achieves the extra objective by an extra multi-head attention:
\begin{equation} 
f_{\text {decoder}}\left(\text{x,y}\right) = \text{FFN}(\text{MHA}(\text{MHA}(\text{x,x,x}),\text{y},\text{y})),
\end{equation}
where $\text{x}$ and $\text{y}$ stand for the query and the key/value, respectively.

Figure \ref{fig:compare} shows a conceptual illustration of the conventional Transformer model, focusing on the attention patterns.
As we can see, it only makes use of the coarse-grained representations at the last encoder layer in decoding, which may fail to capture accurate source information that is better framed by representations from first encoder layers, i.e., fine-grained representations. The effect is more obvious for longer source text (see Sec.~\ref{sec:length}), because it is difficult for the model to find the correct source information in the coarse-grained representations, which can make the generated target sentence unfaithful to the original sentence.

\subsection{Hierarchy Bypassing}
\label{sec:hierarchy_bypassing}

Existing works \cite{Domhan2018How,Bapna2018Training} have tried to narrow the information gap by substituting the representations from the last encoder layer by representations from other encoder layers. 
Taking the generality into consideration, this kind of decoding can be formalized as
\begin{equation}
\text{T}_{i} =f_{\text {decoder}}\left(\text{T}_{i-1}, g_i(\textbf{S})\right), \label{eq:dec-rep}
\end{equation}
where $\textbf{S}=\{\text{{S}}_1, \text{{S}}_2, \ldots, \text{{S}}_{N}\}$ represents the source representations of different granularity from the $N$ encoder layers. The function $g_i(\cdot)$ stands for the layer-wise strategy. For example, supposing the decoding is conducted in a coarse-to-fine manner, that is, the first layer needs high-level information for language planning and the last layer need low-level information for detail realization, one may hypothesize that the decoder layers should draw information from the encoder layers inversely to their positions, i.e., $\text{T}_{i} =f_{\text {decoder}}\left(\text{T}_{i-1}, \text{S}_{N-i+1}\right)$. 
As a result, the first/last decoder layer adopts the information from the last/first encoder layer.

However, such a \textit{direct replacement} may be problematic, leading to the \textit{hierarchy bypassing} problem. To explain the problem, take the previous inversely-connected model as an example. Intuitively, there is a direct connection between the first encoder layer and the last decoder layer shorter than any other connection between the encoder layers and the decoder layers, leading to the problem that the gradient information can pass through the remaining layers of the encoder directly to the first encoder layer. As a result, the representational ability of the encoder is impaired, which in turn leads to performance degradation. 

More specifically, the phenomenon can be explained by the following analysis. We denote the loss function in the conventional model as $\mathcal{E}$ and the $\mathbf{e}_{i}$ and $\mathbf{d}_{i}$ denote the parameters of the $i_{th}$ encoder layer and the $i_{th}$ decoder layer, respectively. We take the most significant $N_{th}$ encoder layer as an example. From the chain rule of backpropagation \cite{LeCun1989Backpropagation}, we can get the gradient of the $N_{th}$ encoder layer in the conventional model and similarly for models with the replacement:
\begin{align}
\left(\frac{\partial \mathcal{E}}{\partial \mathbf{e}_N}\right)_\text{conventional} &= \sum_{i=1}^{N}{\frac{\partial \mathcal{E}}{\partial \mathbf{d}_{i}}\frac{\partial \mathbf{d}_{i}}{\partial \mathbf{e}_{N}}} ,  \\  
\left(\frac{\partial \mathcal{E}}{\partial \mathbf{e}_N}\right)_\text{replaced} &= \frac{\partial \mathcal{E}}{\partial \mathbf{d}_{1}}\frac{\partial \mathbf{d}_{1}}{\partial \mathbf{e}_{N}}.
\end{align}
As we can see, the gradient information of $\left(\nicefrac{\partial \mathcal{E}}{\partial \mathbf{e}_N}\right)_\text{conventional}$ is significantly richer than that of $\left(\nicefrac{\partial \mathcal{E}}{\partial \mathbf{e}_N}\right)_\text{replaced}$.
It means the top encoder layer of the conventional model is better suited to the final output and the top encoder layer of the inversely-connected model is shorted in the sense that its representational ability is not adequately expressed and the representation hierarchy is bypassed in this connection pattern (see Sec.~\ref{sec:Hierarchy_analysis} and Figure~\ref{fig:hierarchy}). This negative effect is confirmed in our experiments (see Sec.~\ref{sec:ablation}) and shown by previous efforts \cite{Domhan2018How}.

\begin{figure*}[t]

\centering
\includegraphics[width=1\linewidth]{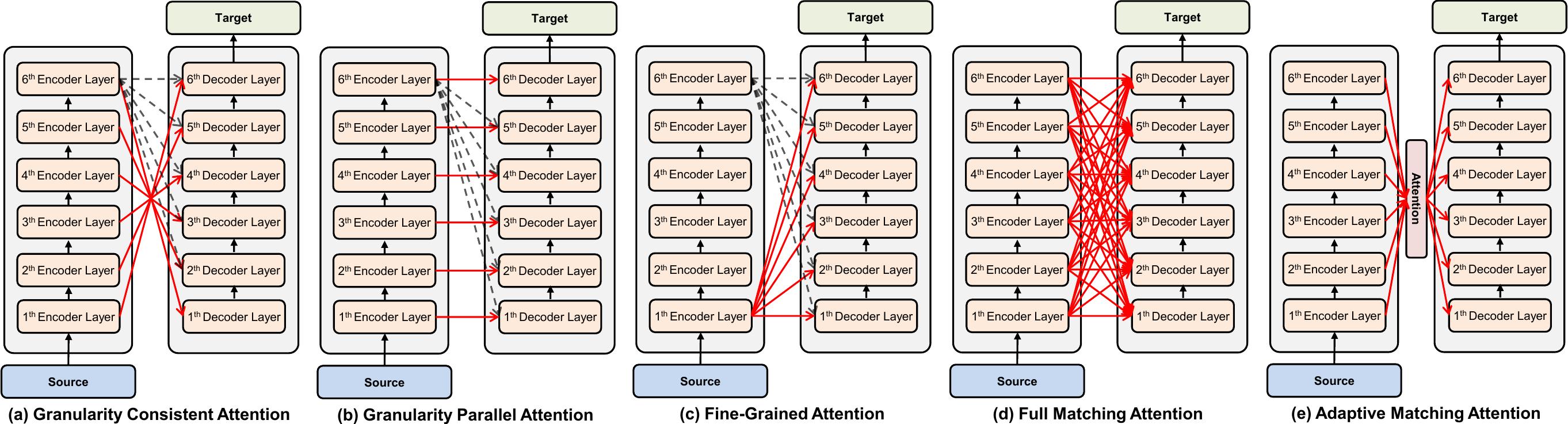}

\caption{We present the proposal on Transformer with various strategies for routing the source representations: (a) Granularity Consistent Attention; (b) Granularity Parallel Attention; (c) Fine-Grained Attention; (d) Full Matching Attention; (e) Adaptive Matching Attention. The dashed lines represent the original attention to the last encoder layer and we omit them in (e) for clarity.} 
\label{fig:model}

\end{figure*}

\subsection{Multi-View Decoding}

To construct a proper way of using deep representations in sequence-to-sequence learning and encourage the decoder to make full use of the source sequence information from the global and local perspectives, we enhance the decoding process with multiple views of the source sequences with different granularity, such that the expressive power of the model could be fully taken advantage of. Particularly, we propose to conduct layer-wise multi-view decoding (see Figure~\ref{fig:model}), for each decoder layer, a different combination of the source views is considered, encouraging the decoder to make adequate, efficient use of the source sequence information from the global and local perspectives. Different from existing work in using deep representations, the focus of our proposal is on crossed views, that is, two different views of the source information is provided at the same time for a decoder layer.

\subsubsection{\textbf{Soft Integration}}
To properly train the encoder stack and address the hierarchy bypassing problem, the final encoder layer needs comprehensive error signals. The simple way to achieve that is to incorporate the additional source views on top of the representations of the final encoder layer. 
The decoder of the final model is updated as follows:
\begin{equation} 
\text{T}_{i} =f_{\text {decoder}}\left(\text{T}_{i-1}, \text{LN}(g_i(\textbf{S})+\text{S}_{N})\right), \label{eq:dec-cross}
\end{equation}
where LN stands for Layer Normalization \cite{Ba2016layernorm}. 
The layer normalization is needed to keep the scale of the representations for stability in deep neural networks and allow the soft integration of two different source views into the model.

\subsubsection{\textbf{Continued Learning}}
For the conventional encoder-decoder model, once trained, its encoder should be capable of describing the source sequence in different granularity. Such ability is crucial to the multi-view decoding. 
However, while applying soft integration alone maintains the richness of gradient sources, the last encoder layers could still underfit due to the co-adaption of the first encoder layer.
Due to the augmentative nature of the proposal, the model structure can be extended seamlessly upon the original model. Thus, we propose to conduct continued training based on the trained conventional model. In practice, the conventional model is first trained normally, and then, our multi-view decoding is applied to further fine-tune the attention structure. Although the structural changes of attention structures pose challenges for continued learning, the model can adapt successfully in reality due to the inherent augmentative design of the approach.

\subsubsection{\textbf{Multi-View Decoding Strategies}}

In the multi-layer setting, each encoder layer adds a level of abstraction and is believed to produce representations that are more and more coarse-grained describing global context. However, in the decoder, the process is more complex, since each decoder layer receives both the source information and target information, which opens up the question of how to properly incorporate different source view, i.e., the function $g_i(\cdot)$ in Eq.~(\ref{eq:dec-rep}) and Eq.~(\ref{eq:dec-cross}). In our work, extending the related literature \cite{Domhan2018How,Bapna2018Training}, we systematically investigate diverse strategies of routing the source views.

\begin{itemize}
\item\smallskip\textbf{Granularity Consistent Attention (GCA):}
Figure~\ref{fig:model} (a) illustrates the connection pattern. 
In this strategy, each decoding step is seen as a realization process, where an abstract, coarse-grained idea turns into concrete, fine-grained words gradually through the layers. Hence, the granularity consistent attention keeps the granularity of source views in order with the decoder layers.
Therefore, the GCA can be defined as $g_i(\textbf{S})=\text{S}_{N-i+1}$.

\item\smallskip\textbf{Granularity Parallel Attention (GPA):}
Opposite to the GCA strategy that keeps the granularity order in attention, the source granularity order in attention is reversed. In this strategy, the decoding is regarded the same as the encoding, where each layer abstracts together the source and the target sequences, hence GPA, defined as $g_i(\textbf{S})=\text{S}_{i}$.
\item\smallskip\textbf{Fine-Grained Attention (FGA):}
Opposite to the conventional Transformer (see Figure~\ref{fig:compare}) that only draws information from the final encoder layer, i.e., coarse-grained representations,
we experiment with adding only the information of the first layer of encoder, i.e., fine-grained representations.
The strategy is named as the Fine-Grained Attention (FGA), which is defined as
$g_i(\textbf{S})=\text{S}_{1}$.

\item\smallskip\textbf{Full Matching Attention (FMA):}
We also consider routing all source views into each decoder layer, i.e., Full Matching Attention (FMA). In implementation, we combine them by linear transformations, and define the FMA strategy as
$g_i(\textbf{S})=\sum_{j=1}^{N}{({W}_{ij}\text{S}_{j}+b_{ij})}$.
Note that, it is layer-wise in that each decoder layer uses different linear transformation parameterized by ${W}_{ij}$ and ${b}_{ij}$. 
\item\smallskip\textbf{Adaptive Matching Attention (AMA):}
Finally, we apply an attention mechanism to help inject information of various granularity levels adaptively, i.e., Adaptive Matching Attention (AMA).
We build an independent vector for each encoder layer to predict the attention weight. We define the AMA as 
$g_i(\textbf{S})=\sum_{j=1}^{N}{{\alpha}_{ij}\text{S}_{j}}$,
where $\alpha_{ij}$ is computed by the attention mechanism and  $\sum_{j=1}^{N}{\alpha}_{ij}=1$. 

\end{itemize}

It should be noted that, in our experiments (please see Section~\ref{sec:analysis_strategies}), we find that all variants are viable in practice and can promote the performance substantially (please see Table~\ref{tab:model_variant}), which validate our motivation and corroborate the effectiveness of the multi-view decoding approach.
However, our experiments also show that GCA performs the best since it could be more in line with the characteristics of the multi-layer encoding and decoding process, while other strategies may route unneeded information, leading to learning difficulties.

\begin{table*}[t]
    \centering
    \caption{Experimental results of machine translation in terms of BLEU. 
    As a whole, the proposed multi-view decoding with granularity consistent attention significantly improves the baselines. More encouragingly, based on the DynamicConv, which is the previous state-of-the-art, our proposal achieves even better performance.
    }
    \label{tab:result-WMT}
    \begin{tabular}{@{}l c l l l@{}}
        \toprule
        Methods   & Year &  EN-DE &  EN-FR   & DE-EN
        \\ \midrule 
        Transformer \cite{Vaswani2017transformer} & 2017 & 28.4  & 41.0 & - \\
        Layer-wise Coordination \cite{He2018Layer-Wise} & 2018  & 29.0 & - & 35.1\\
        Fixup \cite{Zhang2019fixup} & 2019 & 29.3 & -  &  34.5  \\
        Deep Representations \cite{dou2018exploiting} &2018 &  29.2  & - &  -\\
        Fairseq \cite{Ott2018scaling}& 2018  &  29.3  & 43.2 &  - \\
        Evolved Transformer \cite{So2019EvolvedTransformer} & 2019  & 29.8 & 41.3 & - \\
        DynamicConv \cite{wu2019paylessattention} & 2019 & 29.7 &43.2 & 35.2 \\ 
        Anneal Transformer \cite{Lu2021Calibration}
        & 2021 & 28.1 &  - &  -\\
        Transformer+LM+MSO \cite{Miao2021Prevent} & 2021 & 28.6  & 41.7 &  -\\
        \midrule[\heavyrulewidth]
        Transformer (re-implementation) & - &  29.0      & 41.1   &  34.7\\
        \ \ w/ Multi-View Dec. & Ours  & \bf {29.9 (+0.9)} & {42.6 (+1.5)} & {35.9 (+1.2)} \\ \midrule

        DynamicConv (re-implementation)  & - &   29.3   &  43.1   & 35.2  \\ 
        \ \ w/ Multi-View Dec.  & Ours & {29.8 (+0.5)} & \bf {43.5 (+0.4)}   &\bf {36.2 (+1.0)} \\ 
        \bottomrule
    \end{tabular}
\end{table*}

\section{Experiments}
\label{sec:experiment}
Our main experiments focus on neural machine translation \cite{Bahdanau2015seq2seq,Vaswani2017transformer}, which is arguably the most important sequence-to-sequence learning task in natural language processing. We report results using the granularity consistent attention (GCA) for multi-view decoding, which is the best-performed strategy in our preliminary experiments (please see Table~\ref{tab:model_variant}).

\subsection{Datasets}
In our work, we report results on three benchmarks, including two large WMT-2014 datasets\footnote{\url{http://statmt.org/wmt14/}}, i.e., English-German (EN-DE) and English-French (EN-FR), and a small IWSLT-2014 dataset, i.e., German-English (DE-EN).
Follow common practice \cite{Vaswani2017transformer,wu2019paylessattention}, for EN-DE dataset, we use newstest2013 for development and newstest2014 for testing. For EN-FR, we validate on newstest2012+2013 and test on newstest2014. For fair comparisons, following \cite{wu2019paylessattention} and \cite{Vaswani2017transformer}, for WMT EN-DE and EN-FR we measure case-sensitive tokenized BLEU (\texttt{multi-bleu.pl}) against the reference translations. For IWSLT DE-EN, the BLEU is case-insensitive, and since the target language is English, the results are also valid. For WMT EN-DE only, we apply compound splitting similar to \cite{wu2019paylessattention} and \cite{Vaswani2017transformer}. It is unclear how some of the compared methods calculated the BLEU scores, but results provided by us are comparable.

\subsection{Implementation}
As the proposal only relates to the injection of different mix of source representations and is augmentative to the existing models, we keep the inner structure of the baselines untouched and preserve the original settings. For soft integration, we initialize the original attention structures with the parameters of the re-implemented baseline models. For continued learning, we further fine-tune the full model with the number of training steps used to re-implement the baseline model.
We experiment on Transformer \cite{Vaswani2017transformer} and DynamicConv \cite{wu2019paylessattention}. 
Especially, DynamicConv \cite{wu2019paylessattention} established a state-of-the-art in WMT EN-DE and EN-FR translation tasks in comparable settings, i.e., without much larger extra dataset for training as \cite{Edunov2018BackTranslate}. For the re-implementation of the DynamicConv, we use the configuration of six blocks for both encoder and decoder.

Specifically, we use the \textit{fairseq} \cite{ott2019fairseq} for both our re-implementation of baselines and baselines with the proposal. 
For the experiments with Transformer on the two WMT datasets that are much larger, we use the Transformer-Big configuration and train on 8 GPUs. For the experiments on the IWSLT DE-EN dataset, we use the Transformer-Base configuration and train on a single GPU, as it is relatively small. For WMT EN-DE and EN-FR datasets, we also accumulate the gradients for 16 batches before applying an update \cite{Ott2018scaling}, except for Transformer on EN-FR where we do not accumulate gradients.
For all datasets, following in common practice \cite{ott2019fairseq,Ott2018scaling}, we report single model performance by averaging the last 10 checkpoints. Besides, we use beam search of size 4 and length penalty of 0.6 for EN-DE and EN-FR, and use beam search of size 5 for DE-EN.

\subsection{Results}
\begin{itemize}
\item\smallskip\textbf{Automatic Evaluation Results:}
We report results on three benchmarks, including two large WMT-2014 datasets, i.e., English-German (EN-DE) and English-French (EN-FR), and a small IWSLT-2014 dataset, i.e., German-English (DE-EN).
As shown in Table~\ref{tab:result-WMT}, for three datasets, our approach outperforms all the baselines. Based on the Transformer, we promote the baseline by 0.9, 1.5, and 1.2 BLEU scores for the EN-DE, EN-FR, and DE-EN, respectively. More encouragingly, based on the DynamicConv, which is the previous state-of-the-art, our approach sets a new state-of-the-art performance on three datasets, achieving 29.8, 43.5 and 36.2 BLEU scores on EN-DE, EN-FR, and DE-EN respectively. We conduct a significance test on DE-EN and the results are statistically significant (t-test with $p<0.01$).
The improvements on various datasets demonstrate the effectiveness of the proposed approach.

\begin{table}[t]
    \centering
    \caption{Results of human evaluation on the IWSLT DE-EN dataset in terms of faithfulness and fluency.
    \label{tab:human_evaluation}}
    \begin{tabular}{@{}lc c c c@{}}
        \toprule
        Metric  & Loss (\%) & Tie (\%) & Win (\%)   \\ \midrule [\heavyrulewidth]
        
        Faithful & 20 & 34 & \bf 46 \\ 
        
        Fluent  & 18 & 45 & \bf 37 \\ 
        \bottomrule
    \end{tabular}
\end{table}

\begin{table}[t]
    \centering
    \caption{Results of different multi-view strategies on the IWSLT DE-EN dataset in terms of BLEU. 
    }
    \label{tab:model_variant}
    \begin{tabular}{@{}l c c@{}}
        \toprule
        Methods & Transformer & DynamicConv \\  \midrule 
        Baseline & 34.7 & 35.2\\ \midrule 
        Multi-View Dec. w/ GCA & \bf 35.9 & \bf 36.2
        \\  
        Multi-View Dec. w/ GPA  & 35.2 & 35.5\\ 
        Multi-View Dec. w/ FGA & 35.0 &  35.5\\ 
        Multi-View Dec. w/ FMA & 35.4 & 35.7  \\ 
        Multi-View Dec. w/ AMA  & 35.4 & 35.8 \\ 
        \bottomrule
    \end{tabular}
\end{table}

\item\smallskip\textbf{Human Evaluation Results:}
We further conduct a targeted human evaluation in terms of the faithfulness and fluency of the translated sentences.
Specifically, we randomly selected 100 sentences from the IWSLT DE-EN dataset and recruited 5 annotators with sufficient language skills. Each annotator is asked to compare the performance of our approach with the baseline model as Transformer. As shown in Table~\ref{tab:human_evaluation}, our approach enjoys an obvious advantage in terms of the two aspects, meaning the translated sentences contain fewer factual errors and repeated segments. 

\end{itemize}

From the results of automatic and human evaluations, we can see that our proposed multi-view decoding can provide a solid basis for natural language generation.
As a result, our approach can successfully boost baselines and achieves new state-of-the-art results on the EN-DE, EN-FR and DE-EN datasets, which verifies the effectiveness of the proposed approach and indicates that our approach is less prone to the variations of model structures, hyper-parameters (e.g., learning rate and batch size), and learning paradigms.

\section{Analysis}
\label{sec:analysis}
In this section, we conduct systematic analysis from different perspectives to better understand our proposed multi-view decoding approach. 
Unless otherwise specified, we use the Transformer model and the GCA strategy.

\subsection{Multi-View Decoding Strategies}
\label{sec:analysis_strategies}
As reported in Table~\ref{tab:model_variant}, all the strategies can improve the performance and GCA shows the best improvement. 
In comparison, GPA could not provide the fine-grained representations for the last decoder layer; FGA only provides a fixed level of source views; FMA combines all views and may introduce redundant noise, while AMA has a hard time in learning proper weights for different views.
The results demonstrate the GCA strategy could be more in line with the characteristics of the multi-layer encoding and decoding process and the information is provided more efficiently.

\begin{figure}[t]
\centering
\includegraphics[width=0.85\linewidth]{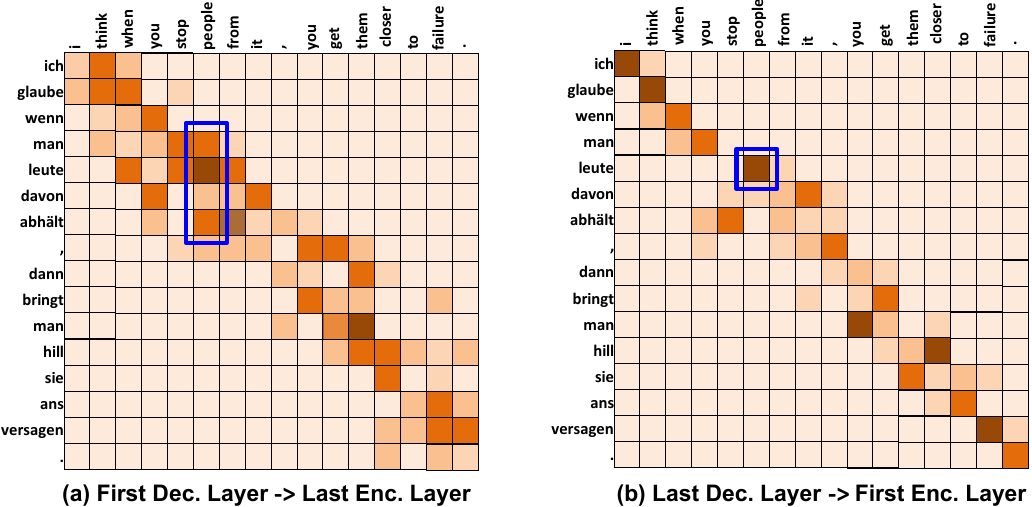}
\caption{Visualization of the attention distribution of the proposed granularity consistent attention (GCA).}
\label{fig:vis}
\end{figure}

\begin{table}[t]
    \centering
    \caption{Ablation results on the IWSLT DE-EN and WMT EN-DE datasets. As we can see, the direct replacement has a negative accuracy impact similarly to \cite{Domhan2018How}, which proves the existence of the hierarchy bypassing problem in the singe-view decoding approaches. Fortunately, both the proposed Soft Integration and Continued Learning can be used to mitigate the hierarchy bypassing problem, significantly improving the performance.}
    \label{tab:ablation_study}
    \begin{tabular}{@{}l c c@{}}
        \toprule
        Methods   & \multicolumn{2}{c}{BLEU}
        \\ \midrule
        Baseline (Transformer)                                            & 34.7 & 29.0 \\  \midrule
        \ \ w/ GCA (Direct Replacement)  & 33.6 & 28.5 \\
        \ \ w/ GCA + Soft Integration   & 34.0 & 29.4 \\ 
        \ \ w/ GCA + Continued Learning  & 34.9 &29.6 \\ \midrule[\heavyrulewidth]
        \ \ \begin{tabular}[c]{@{}l@{}} w/ GCA + Soft Integration + Continued Learning \\ \phantom{w/ GCA +}(i.e., Multi-View Dec.) \end{tabular} & \bf 35.9 & \bf 29.9 \\
        \bottomrule
    \end{tabular}
\end{table}

This explanation can also be supported by the attention distribution instanced in Figure~\ref{fig:vis}, where we show the granularity consistent attention on a sentence. For the first decoder layer attending to the last encoder layer, the attention distribution is fairly dispersed, extracting information for language modeling. For example, as shown in the blue boxes in Figure~\ref{fig:vis}, with the target input word \textit{stop}, the source words with semantic-related roles are attended, i.e., \textit{abh\"alt} ``\textit{to stop}'', \textit{leute} ``\textit{people}'' and \textit{man} ``\textit{you}''. In turn, for the last decoder layer attending to the first encoder layer, the attention is focused directly on the word to be generated, e.g., the source word \textit{leute} when generating \textit{people}.

\begin{figure}[t]
    \centering
    \includegraphics[width=0.6\linewidth]{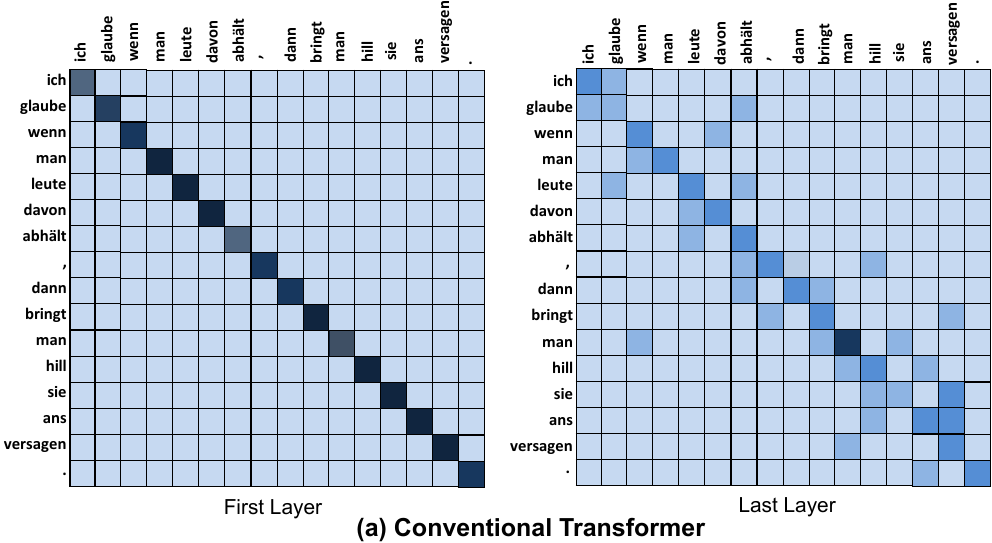}
    \includegraphics[width=0.6\linewidth]{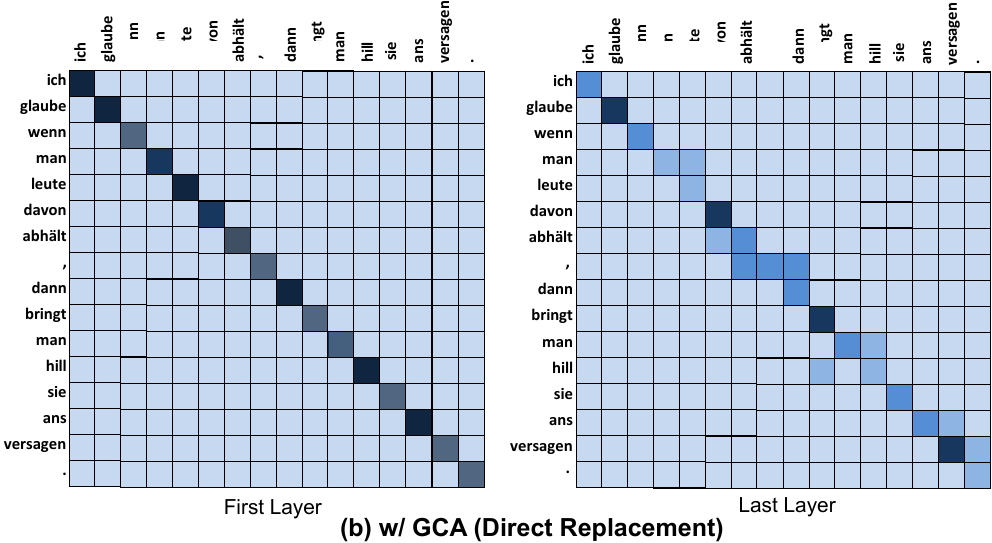}
    \includegraphics[width=0.6\linewidth]{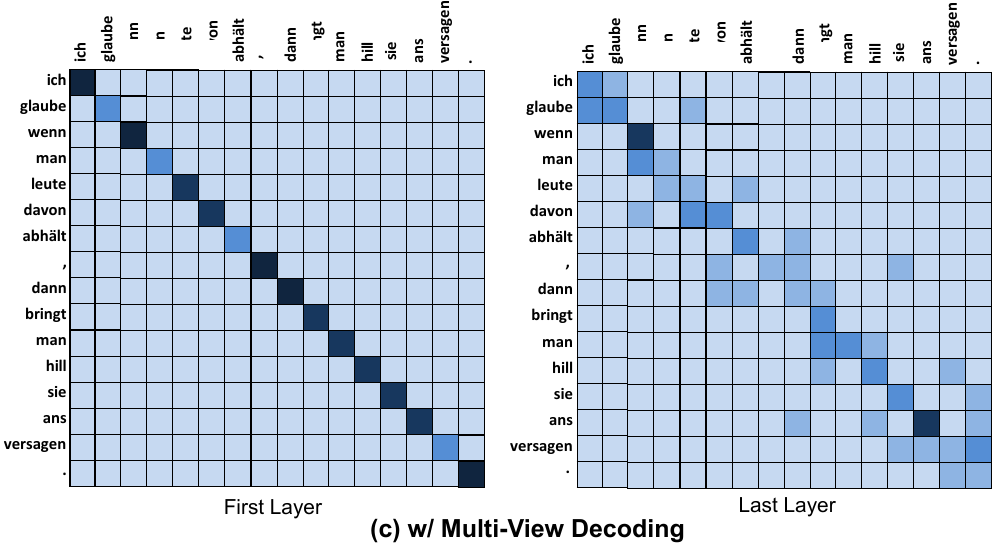}
    \caption{Visualization of the cosine similarity within the source sequence at the first encoder layer (left) and last encoder layer (right). }
    \label{fig:hierarchy}
\end{figure}

\subsection{Ablation Study}
\label{sec:ablation}
In this section, we conduct the ablation study to investigate the contribution of soft integration and continued learning, which are used to mitigate the hierarchy bypassing  problem. The results are reported in Table~\ref{tab:ablation_study}. As we can see, directly replacing the conventional context attention is harmful to the accuracy similar to existing work \cite{Domhan2018How}. If the direct replacement is conducted with continued learning, that is, the model first learns with conventional context attention before GCA, we can observe an improvement that is substantial over the vanilla direct replacement, albeit minimal to the baseline, meaning the single-view approach cannot realize the potential of multi-level source representations. Complementing GCA with the global multi-view has a positive impact, especially when combined with continued learning, reaching an overall improvement of 1.2 BLEU and 0.9 BLEU over the baseline on the DE-EN and EN-DE datasets respectively, indicating the joint advantages of the two proposed methods in addressing the hierarchy bypassing problem.
We also check if the accuracy improvement is simply due to the longer training that comes with continued learning. If the conventional models are allowed the same more training time, their results are not bettered meaningfully. The increase is 0.0, 0.1, and 0.3 in terms of BLEU for DE-EN, EN-DE, and EN-FR, respectively.

\subsection{Hierarchy Bypassing}
\label{sec:Hierarchy_analysis}

The hierarchy bypassing problem can be better understood by analyzing the granularity of the source representations. In Figure~\ref{fig:hierarchy}, we show the cosine similarity within the source sequence of the first and the last encoder layer. For the conventional transformer, i.e., Eq.~(\ref{eq:dec}), we can see that the last encoder layer embodies diffused similarities, indicating more coarse-grained representations, where related information is better associated. However, with direct replacement as in Eq.~(\ref{eq:dec-rep}), the similarities in the last encoder layer remain relatively concentrated, suggesting impaired learning of representation hierarchy and causing the hierarchy bypassing problem. In contrast, with the proposed multi-view decoding, i.e., Eq.~(\ref{eq:dec-cross}), the hierarchy bypassing problem is effectively relieved, enabling the performance improvement of sequence-to-sequence learning with deep hierarchical and diverse representations.

\begin{figure*}[t]

    \centering
    \includegraphics[width=1\linewidth]{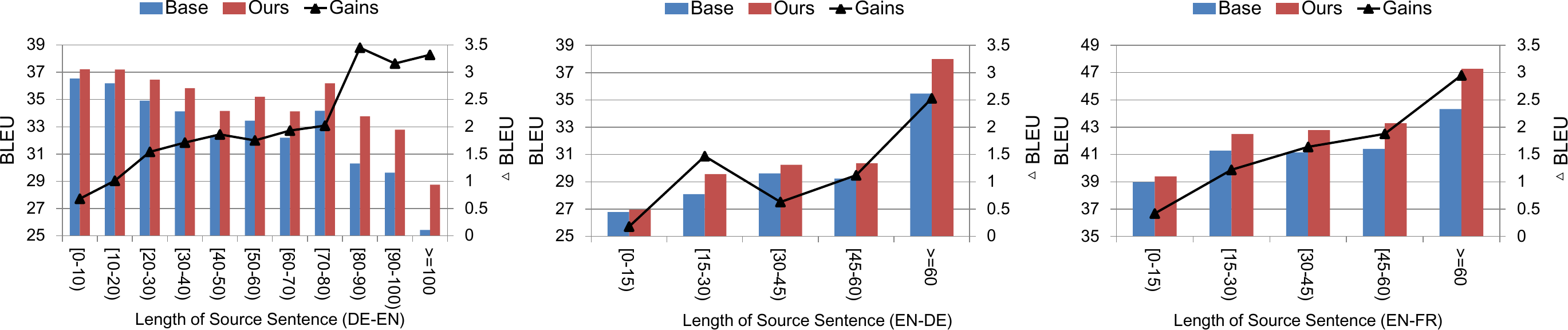}

    \caption{BLEU scores on the test sets with respect to various input sentence lengths. The gains in different length of sentences are shown with the polyline and the right y-axis. }
    \label{fig:length_analysis}
\end{figure*}

\subsection{Length Analysis}
\label{sec:length}
In order to analyze the performance of our approach on sentences of different lengths, following in \cite{Bahdanau2015seq2seq} and \cite{dou2018exploiting}, we group sentences of similar lengths together and calculate the BLEU score for each group. As shown in Figure \ref{fig:length_analysis}, our approach is superior to the baseline in all length segments on DE-EN, EN-DE, and EN-FR datasets. 
It is worth noticing that the proposal is always better than the baseline, and the longer the sentences, the more the improvements. Intuitively, it is hard for the global representation from the final encoder layer to retain all the detailed input information, especially for longer sentences. However, in the conventional encoder-decoder model, the decoder is only equipped with a single view of the source sequence, which causes a dilemma that although both global information and local information are important, only one can be used. In contrast, we can avoid the dilemma by adopting multi-view decoding, which injects fine-grained representations into the decoder and keeps the original global representation at the same time.

\subsection{Generalization Analysis}
It is interesting to see whether our approach works for other sequence-to-sequence tasks. To this end, based on GCA, we further conduct experiments on the CNN-Daily Mail dataset  \cite{Hermann2015Teaching} for abstractive summarization, the COCO dataset  \cite{chen2015microsoft} for image captioning, the MSVD dataset \cite{Guadarrama2013MSVD} and the MSR-VTT \cite{Xu2016MSR-VTT} dataset for video captioning, the MIMIC-CXR dataset \cite{Johnson2019MIMIC} and the IU-Xray dataset \cite{Dina2016IU-Xray} for medical report generation, and the Twitter dataset \cite{Lan2017twitter} and the Quora dataset for paraphrase generation. The improvements are all statistically significant (t-test with $p<0.01$).

\subsubsection{\textbf{Abstractive Summarization}}
The abstractive summarization task is able to test the ability of our approach to deal with longer texts.

\begin{itemize}
\item\smallskip\textbf{Dataset:}
We train the models on the CNN-Daily Mail dataset \cite{Hermann2015Teaching}, which contains online news articles (781 tokens on average) paired with multi-sentence summaries (56 tokens on average). 
Following \cite{see2017Point}, we truncate each source sentence to 400 words and each target sentence to 100 words.  ROUGE-1, -2 and -L \cite{lin2004rouge} are used to evaluate the performance of models. The dataset is able to test the ability of our approach to deal with longer texts.

\item\smallskip\textbf{Implementation:}
We use the default setting provided by \textit{OpenNMT} \cite{opennmt}. For the experiments with Transformer, we use the Transformer-Base configuration and train on a single GPU. 
When generating summaries, we follow standard practice in tuning the maximum output length, disallowing repeating the same trigram, and applying a stepwise length penalty \cite{Paulus2018Deep,Fan2018Controllable}. 

\item\smallskip\textbf{Results:}
Table~\ref{tab:abstractive} shows the results in terms of ROUGE-1, -2, and -L \cite{lin2004rouge}. The proposal achieves an advantage over the baseline. It indicates that our approach generalizes well to tasks with much longer source sequences, which are around 400 words, and is effective in dealing with longer sequences. 
In fact, in such scenarios, where the summary should be concise but also accurate in detail, the GCA naturally streamlines the decoding process.

\end{itemize}

\subsubsection{\textbf{Image Captioning}}
The task of image captioning aims to generate a textual description for an input image, thus this task combines image understanding and language generation and is a cross-modal setting compared to machine translation and abstractive summarization. The image captioning task belongs to the sequence-to-sequence task, where the source sequence and target sequence contain non-ordered region-of-interest features \cite{anderson2018bottom} and generated captions, respectively.

\begin{table}[t]
    \centering
    \caption{Results of abstractive summarization on CNN-Daily Mail dataset \cite{Hermann2015Teaching}. We conducted 5 runs with different seeds for all experiments, the \ssymbol{3} denotes statistically significant results (t-test with $p<0.01$). As we can see, our approach achieves the best results on the CNN-Daily Mail abstractive summarization dataset across all metrics.
    \label{tab:abstractive}}
    \setlength{\tabcolsep}{1.5pt}
    \begin{tabular}{@{}l l l l@{}}
        \toprule
        \multirow{2}{*}[-3pt]{Methods} &  \multicolumn{3}{c}{Abstractive Summarization}  \\ \cmidrule(r){2-4} & ROUGE-1  & ROUGE-2 & ROUGE-L  \\ \midrule 
        LSTM \cite{Paulus2018Abstractive} &38.3 &14.8 &35.5 \\
        CNN \cite{Fan2018Abstractive} &39.1 &15.4 &35.8\\
        LightConv \cite{wu2019paylessattention} & 39.5 & 16.0 & 36.5\\ 
        DynamicConv \cite{wu2019paylessattention} & 39.8 & 16.3 &36.7 \\ \midrule 
        Transformer(re-implementation)  & 39.3  &  17.3  & 36.2 \\
        \ \ w/ Multi-View Dec.  &  \bf{39.9 (+0.6)}\ssymbol{3}   &  \bf{18.0 (+0.7)}\ssymbol{3} &  \bf{36.8 (+0.6)}\ssymbol{3} \\ 
        \bottomrule
    \end{tabular}
\end{table}

\begin{itemize}
\item\smallskip\textbf{Dataset:}
We conduct experiments on the popular COCO dataset \cite{chen2015microsoft}, which contains 123,287 images, each of which is is annotated with 5 sentences. 
Following common practice \cite{anderson2018bottom,Yao2018exploring,Yang2019Auto-Encoding,Zhang2021RSTNet}, we use the publicly-available splits in \cite{karpathy2014deep} for offline evaluation. There are 5,000 images each in validation set and test set. Following common practice, we replace caption words that occur less than 5 times in the training set with the generic unknown word token UNK, resulting in a vocabulary of 9,487 words. We report results with the help of the evaluation toolkit  \cite{chen2015microsoft}, which includes the commonly-used metrics SPICE \cite{anderson2016spice} and CIDEr \cite{vedantam2015cider}, which are customized for evaluating image captioning systems, based on scene-graph and n-gram matching, respectively, which are more consistent with human judgment \cite{anderson2016spice,Vinyals2017Lessons}.

\begin{table}[t]
    \centering
    \caption{Results of image captioning on COCO dataset \cite{chen2015microsoft}. We conducted 5 runs with different seeds for all experiments, the \ssymbol{3} denotes statistically significant results (t-test with $p<0.01$). As we can see, our approach achieves the best results on the COCO image captioning dataset across all metrics.
    \label{tab:imagecaption}}
    \setlength{\tabcolsep}{4pt}
    \begin{tabular}{@{}l c l l@{}}
        \toprule
        \multirow{2}{*}[-3pt]{Methods} & \multirow{2}{*}[-3pt]{Year} & \multicolumn{2}{c}{Image Captioning}  \\ \cmidrule(r){3-4} & & SPICE & CIDEr \\ \midrule 
        Up-Down \cite{anderson2018bottom} & 2018 & 21.4 & 120.1 \\
        ORT \cite{Herdade2019ORT} & 2019 & 22.6 &128.3  \\
        GLIED \cite{liu2019GLIED} & 2019 & 22.6 & 129.3  \\
        Prophet \cite{liu2020prophet} & 2020 & 23.3 & 133.4 \\
        X-Transformer \cite{Pan2020XLAN} & 2020 &23.4 &132.8 \\
        RSTNet \cite{Zhang2021RSTNet} & 2021 & 23.3 & 135.6 \\ \midrule
        Transformer (re-implementation)  & - & 21.2  & 124.9\\
        \ \ w/ Multi-View Dec.  & Ours & \bf{22.6 (+1.4)}\ssymbol{3} & \bf {129.4 (+4.5)}\ssymbol{3}\\  \midrule
        
        X-Transformer (re-implementation) & - & 23.1  & 132.7 \\
        \ \ w/ Multi-View Dec.  & Ours & \bf{24.1 (+1.0)}\ssymbol{3} & \bf {136.5 (+3.8)}\ssymbol{3}\\  
        
        \bottomrule
    \end{tabular}
\end{table}
\begin{table*}[t]
\centering
\renewcommand\tabcolsep{4pt}
\caption{Leaderboard performance on the online MSCOCO image captioning evaluation server, where the ground truth captions are not available. c5 means comparing to 5 references and c40 means comparing to 40 references. As we can see, our approach outperforms all current published state-of-the-art models across all metrics over the online leaderboard, which further prove the effectiveness of our proposed approach.
\label{tab:res-server}}
\begin{tabular}{@{}l c c c c c c c c c c c c c c c@{}}
\toprule
\multirow{2}{*}[-3pt]{Methods}  
& \multicolumn{2}{c}{BLEU-1}
&  \multicolumn{2}{c}{BLEU-2}
&  \multicolumn{2}{c}{BLEU-3}
&  \multicolumn{2}{c}{BLEU-4}
&  \multicolumn{2}{c}{METEOR}
& \multicolumn{2}{c}{ROUGE-L}
& \multicolumn{2}{c}{CIDEr}  \\
\cmidrule(lr){2-3} \cmidrule(lr){4-5} \cmidrule(lr){6-7} \cmidrule(lr){8-9} \cmidrule(lr){10-11} \cmidrule(lr){12-13} \cmidrule(lr){14-15}

& c5 & c40 & c5 & c40 & c5 & c40 & c5 & c40 & c5 & c40 & c5 & c40 & c5 & c40 \\
\midrule

SCST   \cite{rennie2017self}            &78.1           & 93.7     & 61.9     & 86.0     & 47.0 &75.9 &35.2 &64.5 &27.0 &35.5 &56.3 &70.7 &114.7 &116.7     \\
Up-Down \cite{anderson2018bottom}  & 80.2     & 95.2 &64.1     & 88.8     &  49.1 &79.4 &36.9 &68.5 &27.6 &36.7 &57.1 &72.4 &117.9 &120.5 \\
CAVP \cite{liu2018context}  & 80.1     &  94.9 &64.7     & 88.8     &  50.0 &79.7 &37.9 &69.0 &28.1 &37.0 &58.2 &73.1 &121.6 &123.8 \\
ETA \cite{Li2019ETA} &81.2 &95.0 &65.5 &89.0 &50.9 &80.4 &38.9 &70.2 &28.6 &38.0 &58.6 &73.9 &122.1 &124.4 \\
RFNet \cite{jiang2018recurrent}   & 80.4     & 95.0 & 64.9 & 89.3 &  50.1 &80.1 &38.0 &69.2 &28.2 &37.2 &58.2 &73.1 &122.9 &125.1 \\
GLIED \cite{liu2019GLIED} &80.1 &94.6 &64.7 &88.9 &50.2 &80.4 &38.5 &70.3 &28.6 &37.9 &58.3 &73.8 &123.3 &125.6 \\
GCN-LSTM \cite{Yao2018exploring} & - & - &65.5 &89.3 &50.8 &80.3 &38.7 &69.7 &28.5 &37.6 &58.5 &73.4 &125.3 &126.5 \\
SGAE \cite{Yang2019Auto-Encoding} &81.0 &95.3 &65.6 &89.5 &50.7 &80.4 &38.5 &69.7 &28.2 &37.2 &58.6 &73.6 &123.8 &126.5 \\
$\mathcal{M}^{2}$ Trans. \cite{Cornia2020M2} & 81.6 & 96.0 & 66.4 & 90.8 & 51.8 & 82.7 & 39.7 & 72.8 & 29.4 & 39.0 & 59.2 & 74.8 & 129.3 & 132.1 \\
X-LAN \cite{Pan2020XLAN} & 81.1 & 95.3 & 66.0 & 89.8 & 51.5 & 81.5 & 39.5 & 71.4 & 29.4 & 38.9 & 59.2&  74.7&  128.0 & 130.3 \\
X-Transformer \cite{Pan2020XLAN} &81.3 &95.4& 66.3 &90.0 & \bf51.9& 81.7 & \bf39.9& 71.8 &29.5& 39.0 &59.3& 74.9 &129.3 &131.4 \\
RSTNet \cite{Zhang2021RSTNet}&  \bf 81.7 &  \bf96.2 &  \bf66.5 & 90.9 & 51.8 & 82.7 & 39.7 & 72.5&  29.3 & 38.7 & 59.2 & 74.2 & 130.1 & 132.4 \\
\midrule
Ours & \bf 81.7& \bf 96.2&	66.4&\bf	91.1&	51.8&	 \bf 83.1&	39.7&	 \bf73.2& \bf	29.6& \bf	39.3&	 \bf59.4& \bf	75.0& \bf	130.2& \bf	133.5 \\
\bottomrule
\end{tabular}
\vspace{-10pt}
\end{table*}

\item\smallskip\textbf{Implementation:}
For experiments on the image captioning dataset, we use the Transformer-base model and train on a single GPU. 
For fair comparisons, we use the RCNN-based image features provided by \cite{anderson2018bottom}. 
We train the model with both cross-entropy loss and reinforcement learning optimizing CIDEr. The model is trained with batch size of 50 for 25 epochs with early stopping based on CIDEr with cross-entropy loss, followed by reinforcement learning. 
We use Adam with a learning rate of $10^{-4}$ for parameter optimization. 
During inference, we apply beam search with beam size = 5.

\item\smallskip\textbf{Offline Evaluation Results:}
Table~\ref{tab:imagecaption} reports the results on the test set in terms of SPICE \cite{anderson2016spice} and CIDEr \cite{vedantam2015cider}, which are specifically designed to evaluate image captioning systems.
As we can see, for the image captioning task, all baselines equipped with our approach receive performance gains over all metrics.
Specifically, our approach further improves the performance of baseline Transformer to 22.6 SPICE score and 129.4 CIDEr score, and improves the performance of baseline X-Transformer to 24.1 SPICE score and 136.5 CIDEr score which outperforms the current state-of-the-art model RSTNet \cite{Zhang2021RSTNet}.
The results suggest that the proposed approach could be extended to a wide range of sequence generation tasks with various kinds of source representations, demonstrating the universality of our approach. 

\item\smallskip\textbf{Online Evaluation Results:}
Following common practice \cite{anderson2018bottom,Yao2018exploring,Yang2019Auto-Encoding,Zhang2021RSTNet}, we also evaluate our approach on the online MSCOCO evaluation server\footnote{\url{https://competitions.codalab.org/competitions/3221\#results}}, where the ground truth captions are not available.
For online evaluation, all of the recently submitted systems use model ensemble method \cite{anderson2018bottom,Yang2019Auto-Encoding,Yao2018exploring}.
Therefore, following existing works, we submit an ensemble of four ``X-Transformer w/ Multi-View Dec.'' models to the leaderboard and compare with the top-performing entries on the leaderboard whose methods are published, which including RSTNet \cite{Zhang2021RSTNet}, X-Transformer \cite{Pan2020XLAN}, SGAE \cite{Yang2019Auto-Encoding}.
From Table~\ref{tab:res-server}, we can find that our approach is able to achieve new state-of-the-art results in major metrics over the leaderboard, which further verifies the effectiveness of our proposed \textit{layer-wise multi-view decoding}.

\end{itemize}

\subsubsection{\textbf{Video Captioning}}
Video captioning targets to understand the visual content of given videos and generate corresponding descriptive sentences. Compared with image captioning, the video captioning task is relatively more challenging, because the video involves various scenes that are volatile and likely to change. Besides, there are three source modalities (i.e., image, motion, and audio) and the temporal dynamics information should be captured to understand the video efficiently, while image captioning only use image as input. Therefore, injecting fine-grained representations of the source sequence, i.e., a sequence of frames, is helpful to generate accurate video captions.

\begin{itemize}
\item\smallskip\textbf{Dataset:}
We evaluate the performance of our approach on the benchmark Microsoft Video Description (MSR-VTT) \cite{Xu2016MSR-VTT} dataset and Microsoft Video Description (MSVD) \cite{Guadarrama2013MSVD} dataset.
In detail, MSR-VTT contains 10,000 video clips, and each video is annotated with 20 annotated sentences.
Following \cite{Pei2019MARN,Pan2020Spatio}, we adopt the official splits to report our results, resulting in there being 6513, 497, and 2990 video clips in the training, validation, and test sets, respectively.
MSVD includes 1,970 video clips and roughly 80,000 English sentences. We follow the split settings in \cite{Pei2019MARN}, resulting in 1,200, 100, and 670 videos for the training, validation, and test sets, respectively.
We replace caption words that occur less than 3 times in the training set with the [UNK] token, plus with a [MASK] token, resulting in a vocabulary of 10,546 words for MSR-VTT and 9,467 words for MSVD.

\begin{table*}[t]
\centering
\setlength{\tabcolsep}{2.5pt}
\caption{Results of video captioning on MSVD and MSR-VTT datasets.
The \ssymbol{3} denotes statistically significant results (t-test with $p<0.01$).
As we can see, our approach can significantly boost the performance of the baseline model and set new state-of-the-arts across major metrics on video captioning.}
\label{tab:videocaptioning}
\begin{tabular}{@{}l c l l l l@{}}
\toprule 

\multirow{2}{*}[-3pt]{Methods} & \multirow{2}{*}[-3pt]{Year} & \multicolumn{4}{c}{Video Captioning Dataset: MSVD \cite{Guadarrama2013MSVD}}  \\ \cmidrule(lr){3-6}  
& & BLEU-4 & METEOR & ROUGE-L & CIDEr  \\
\midrule [\heavyrulewidth]

MARN \cite{Pei2019MARN} & 2019 & 48.6 & 35.1&  71.9 & 92.2  \\
GRU-EVE \cite{Aafaq2019GRU-EVE} & 2019 & 47.9 & 35.0 & 71.5 & 78.1  \\
POS-Control \cite{Wang2019Controllable} & 2019 & 52.5&  34.1 & 71.3&  88.7   \\
STAT \cite{Yan2020STAT}& 2020  & 52.0 & 33.3 & - & 73.8  \\
STGN-OAKD \cite{Pan2020Spatio}& 2020  & 52.2&  36.9&  73.9 & 93.0  \\
ORG-TRL \cite{Zhang2020ORG} & 2020 & 54.3 & 36.4 & 73.9&  95.2 \\
SAAT \cite{Zheng2020SAAT} & 2020 & 46.5 & 33.5&  69.4 & 81.0 \\
SGN \cite{Ryu2021SGN} & 2021 &  52.8 & 35.5 & 72.9 & 94.3    \\
\midrule 

Transformer (re-implementation) & - & 49.8 & 35.1 & 72.2 & 91.2   \\
\ \ w/ Multi-View Dec. & Ours  & \bf{53.4 (+3.6)}\ssymbol{3} & \bf {36.7 (+1.6)}\ssymbol{3}  & \bf {74.6 (+2.4)}\ssymbol{3} & \bf {97.4 (+6.2)}\ssymbol{3}   \\ 

\midrule [\heavyrulewidth]

\multirow{2}{*}[-3pt]{Methods} & \multirow{2}{*}[-3pt]{Year}   & \multicolumn{4}{c}{Video Captioning Dataset: MSR-VTT \cite{Xu2016MSR-VTT}} \\ \cmidrule(lr){3-6}  
& & BLEU-4 & METEOR & ROUGE-L & CIDEr \\
\midrule [\heavyrulewidth]

MARN \cite{Pei2019MARN} & 2019  & 40.4 & 28.1 & 60.7 & 47.1\\
GRU-EVE \cite{Aafaq2019GRU-EVE} & 2019  &  38.3&  28.4 & 60.7 & 48.1 \\
POS-Control \cite{Wang2019Controllable} & 2019   & 42.0 & 28.2&  61.6&  48.7 \\
STAT \cite{Yan2020STAT}& 2020   & 39.3 & 27.1 & - & 43.8 \\
STGN-OAKD \cite{Pan2020Spatio}& 2020    & 40.5 & 28.3&  60.9 & 47.1  \\
ORG-TRL \cite{Zhang2020ORG} & 2020   &  43.6 &  28.8&  62.1 & 50.9 \\
SAAT \cite{Zheng2020SAAT} & 2020  &  39.9 & 27.7 & 61.2 & 51.0\\
SGN \cite{Ryu2021SGN} & 2021   & 40.8&  28.3 & 60.8 & 49.5  \\
\midrule 

Transformer (re-implementation) & -   & 43.2 & 28.5 & 61.8 & 49.3  \\
\ \ w/ Multi-View Dec. & Ours    &  \bf {44.7 (+1.5)}\ssymbol{3} & \bf {29.7 (+1.2)}\ssymbol{3} & \bf {63.5 (+1.7)}\ssymbol{3} & \bf {53.2 (+3.9)}\ssymbol{3} \\ 

\bottomrule
\end{tabular}
\end{table*}
\begin{table*}[t]
\centering
\caption{Results of medical report generation on MIMIC-CXR and IU-Xray datasets. Similarly, our approach brings significant improvements to the baseline model Transformer and surpasses the previous state-of-the-art methods across all metrics on the MIMIC-CXR and IU-Xray datasets.
\label{tab:report_generation}}
\begin{tabular}{@{}l c l l l l l@{}}
\toprule

\multirow{2}{*}[-3pt]{Methods} & \multirow{2}{*}[-3pt]{Year} & \multicolumn{5}{c}{Medical Report Generation Dataset: MIMIC-CXR \cite{Johnson2019MIMIC}} \\ \cmidrule(lr){3-7} & & BLEU-1 & BLEU-4 & METEOR & ROUGE-L & CIDEr \\
\midrule
CNN-RNN \cite{Vinyals2015Show} & 2015 &29.9 &8.4 &12.4 &26.3 & -  \\
AdaAtt \cite{lu2017knowing} & 2017 & 29.9 &8.8 &11.8 &26.6 & - \\
Att2in \cite{rennie2017self} & 2017 &32.5 &9.6 &13.4 &27.6 & - \\ 
Up-Down \cite{anderson2018bottom} & 2018 & 31.7  &9.2 &12.8 &26.7  & - \\
R2Gen \cite{Chen2020Generating} & 2020 &35.3 &10.3 &14.2 &27.7 & - \\ 
PPKED \cite{liu2021PPKED} & 2021 &  36.0  &  10.6 &  14.9 &  28.4 &  23.7  \\ 
\midrule

Transformer (re-implementation) & - & 33.4  & 10.2 & 13.5 & 27.1 & 19.6 \\
\ \ w/ Multi-View Dec.   & Ours & \bf {37.1 (+3.7)}\ssymbol{3} & \bf {11.0 (+0.8)}\ssymbol{3}& \bf {15.4 (+1.9)}\ssymbol{3}& \bf {29.3 (+2.2)}\ssymbol{3}& \bf {25.5 (+5.9)}\ssymbol{3}
\\
\midrule [\heavyrulewidth]

\multirow{2}{*}[-3pt]{Methods} & \multirow{2}{*}[-3pt]{Year} & \multicolumn{5}{c}{Medical Report Generation Dataset: IU-Xray \cite{Dina2016IU-Xray}} \\ \cmidrule(lr){3-7} & & BLEU-1 & BLEU-4 & METEOR & ROUGE-L & CIDEr \\ \midrule
CoAtt \cite{Jing2018Automatic} & 2018 & 45.5  & 15.4 & - & 36.9 & 27.7\\
HRGR-Agent \cite{Li2018Hybrid} & 2018  &43.8  & 15.1 & - & 32.2 & 34.3\\
CMAS-RL \cite{Jing2019Show} & 2019 &46.4  &15.4 & - &36.2  &27.5 \\
SentSAT+KG \cite{Zhang2020When} & 2020 &44.1  &14.7 & - &36.7 & 30.4 \\
R2Gen \cite{Chen2020Generating} & 2020 &47.0 &16.5 & 18.7 & 37.1  & - \\
PPKED \cite{liu2021PPKED} & 2021 &  48.3 &  16.8 &  19.0 &  37.6 &  35.1 
\\  \midrule
Transformer (re-implementation) & - &  46.7 & 15.7 & 18.4 & 35.9 & 33.5 \\
\ \ w/ Multi-View Dec. & Ours  & \bf {48.5 (+1.8)}\ssymbol{3} & \bf {16.9 (+1.2)}\ssymbol{3} & \bf {19.6 (+1.2)}\ssymbol{3} & \bf {38.0 (+2.1)}\ssymbol{3} & \bf {38.3 (+4.8)}\ssymbol{3}\\
\bottomrule
\end{tabular}
\end{table*}

\item\smallskip\textbf{Implementation:} In this experiment, we adopt the Transformer-base model as our baseline model and train on a single GPU.
For fair comparisons \cite{Pan2020Spatio}, given a video, $N=8$ key frames are uniformly sampled to extract image features via the Inception-ResNet-V2 \cite{Szegedy2017Inception-v4} pre-trained on the ImageNet \cite{Deng2009ImageNet}. Considering both the past and the future contexts, we take each key frame as the center to generate corresponding motion features and audio features.
In detail, we adopt the C3D network \cite{Tran2015C3D} pre-trained on Sports-1M dataset \cite{Karpathy2014sports-1M} to extract the motion features and adopt the Bag-of-Audio-Words (BoAW) \cite{Pancoast2014boaw}, Fisher Vector \cite{Jorge2013FisherVector} and VGGish \cite{Hershey2017VGGish} to extract the audio features.
During inference, we also apply beam  search with beam size = 3.

\item\smallskip\textbf{Results:}
As shown in Table~\ref{tab:videocaptioning}, our multi-view decoding approach can successfully boost the baseline model Transformer, with the most significant improvement up to relatively {7\%} and {8\%} for the MSVD dataset and MSR-VTT dataset in terms of CIDEr, respectively. The CIDEr is specifically designed to evaluate captioning systems \cite{vedantam2015cider}.
More encouragingly, our approach can achieve new state-of-the-art results on the MSVD and MSR-VTT video captioning datasets, respectively. 
It is worth noticing that the POS-Control \cite{Wang2019Controllable}, GRU-EVE \cite{Aafaq2019GRU-EVE}, and SAAT \cite{Zheng2020SAAT} adopt the same features as our approach,
the significant improvements brought by our approach using the same features further demonstrate the effectiveness and generalization ability of our proposed multi-view decoding approach.

\end{itemize}

\subsubsection{\textbf{Medical Report Generation}}
Different from the image captioning task, the medical report generation task aims to generate long and coherent descriptions of the input medical images. Thus, this experiment can verify the effectiveness of our approach in dealing with low-resource and long biomedical texts.

\begin{itemize}
\item\smallskip\textbf{Dataset:}
We conduct experiments on two public benchmark datasets,
i.e., IU-Xray \cite{Dina2016IU-Xray} and MIMIC-CXR \cite{Johnson2019MIMIC}, where the former contains 7,470 chest X-ray images associated with 3,955 radiology reports and the latter contains 377,110 chest X-ray images associated with 227,835 radiology reports.
For the IU-Xray dataset, following common practice \cite{Chen2020Generating,Li2019Knowledge,Li2019Knowledge}, we randomly split the dataset into 70\%-
10\%-20\% training-validation-testing splits. 
For MIMIC-CXR, following \cite{Chen2020Generating,liu2021PPKED,Chen2021Cross}, we adopt the official splits to report our results, acquiring 368,960/2,991/5,159 samples in the training/validation/test sets. At last, we convert all tokens of reports to lower cases and filter tokens that occur less than 10 times in the corpus, resulting in a vocabulary of around 4,000 tokens. We adopt the BLEU, METEOR, ROUGE-L, and CIDEr to evaluate the performance.

\item\smallskip\textbf{Implementation:}
We choose the Transformer-base model as our baseline model.
To re-implement the baseline Transformer model and our approach, following common practice \cite{Jing2019Show,Li2019Knowledge,Li2018Hybrid,liu2021PPKED}, we adopt the ResNet-50 \cite{he2016deep}, which is pre-trained on the ImageNet dataset \cite{Deng2009ImageNet} and fine-tuned on CheXpert dataset \cite{Irvin2019CheXpert}, to extract the image features.

\item\smallskip\textbf{Results:}
The results on the MIMIC-CXR and IU-Xray medical report generation datasets are reported in Table~\ref{tab:report_generation}.
As we can see, our multi-view decoding approach can substantially boost the baseline Transformer model.
Specifically, on the MIMIC-CXR dataset, the CIDEr score can be improved by 30\%.
Similarly, on the IU-Xray dataset, the BLEU-4 score and CIDEr score can be improved by 8\% and 14\%, respectively.
Moreover, we compare our approach with several competitive models including the current state-of-the-art models, i.e., SentSAT + KG \cite{Zhang2020When}, R2Gen \cite{Chen2020Generating}, and PPKED \cite{liu2021PPKED}, whose results are directly copied from their original paper.
As we can see, our approach can outperform these existing state-of-the-art models across all metrics on the IU-Xray and MIMIC-CXR datasets, which further corroborates the effectiveness and generalization capabilities of our proposed approach to a wide range of sequence-to-sequence tasks.

\end{itemize}

\begin{table}[t]
    \centering
    \caption{Results of paraphrase generation on Twitter dataset and Quora dataset. As we can see, our approach achieves the best results on two paraphrase generation benchmark datasets across all metrics.
    \label{tab:paraphrase}}
    \setlength{\tabcolsep}{1pt}
    \begin{tabular}{@{}l c l l l l l l@{}}
        \toprule
        \multirow{2}{*}[-3pt]{Methods} & \multirow{2}{*}[-3pt]{Year} & \multicolumn{3}{c}{
        Paraphrase Generation: Quora} & \multicolumn{3}{c}{Paraphrase Generation: Twitter}  \\ \cmidrule(r){3-5} \cmidrule(r){6-8} & & ROUGE-2 & BLEU & METEOR & ROUGE-2 & BLEU & METEOR  \\ \midrule  
        Seq2Seq \cite{Bahdanau2015seq2seq} & 2015 & 35.87 &27.51 &29.45 & 18.23& 13.21& 15.33 \\
        Pointer \cite{see2017Point} & 2017 & 37.28& 28.25& 30.90 &22.20&16.32& 18.18 \\
        Residual LSTM \cite{Prakash2018Paraphrase} & 2018 &37.20& 28.76& 30.44& 20.51& 14.61 &16.53  \\
        RL-ROUGE \cite{Yuanxin2019Paraphrase} & 2018 &37.33& -& 30.96&22.99 &- &18.89 \\
        RbM-SL \cite{li2018Paraphrase} & 2018 & 38.11 &- &32.84 &24.23 &-& 19.97 \\ 
        Prophet \cite{liu2020prophet} & 2020 & 41.24 &32.36 &33.18 &26.83 &18.89 & 21.04
        
        \\ \midrule
        Transformer  & - & 40.14 & 31.16 & 31.68  & 22.67 & 16.19 & 17.63 \\
        \ \ w/ Multi-View Dec.  & Ours & \bf{43.57 (+3.43)}\ssymbol{3} & \bf {33.49 (+2.33)}\ssymbol{3} & \bf {35.36 (+3.68)}\ssymbol{3} & \bf {27.32 (+4.65)}\ssymbol{3} & \bf {19.37 (+3.18)}\ssymbol{3} & \bf {21.91 (+4.28)}\ssymbol{3}\\

        \bottomrule
    \end{tabular}
\end{table}

\subsubsection{\textbf{Paraphrase Generation}}
Paraphrases convey the same meaning as the original sentences or text, but with different expressions in the same language (i.e., with variations in lexicon or syntax). Paraphrase generation aims to synthesize paraphrases of a given sentence automatically \cite{Madnani2010Survey,Passonneau2018Wise,Yuanxin2019Paraphrase}. 
This is a fundamental natural language processing task, and it is important for many downstream applications \cite{Madnani2010Survey,Passonneau2018Wise}, e.g., query rewriting \cite{dong2017Paraphrase}, data augmentation \cite{Iyyer2018Adversarial} and language model pre-training \cite{Lewis2020Paraphrasing}.

\begin{itemize}
\item\smallskip\textbf{Dataset:}
Following common practice, we adopt two benchmark datasets, i.e., Twitter dataset \cite{Lan2017twitter} and Quora dataset\footnote{\url{https://data.quora.com/First-Quora-Dataset-Release-Question-Pairs}}, and the widely-used automatic evaluation metircs: BLEU \cite{papineni2002bleu}, ROUGE \cite{lin2004rouge}, and METEOR \cite{banerjee2005meteor}, to evaluate the performance of models.

\item\smallskip\textbf{Implementation:}
For the experiments, we implement the standard Transformer-BASE model \cite{Vaswani2017transformer} as our baseline.
We use the default setting provided by \textit{OpenNMT} \cite{opennmt}. For the experiments with Transformer, we use the Transformer-Base configuration and train on a single GPU.

\item\smallskip\textbf{Results:}
In Table~\ref{tab:paraphrase}, as we can see, our approach can improve the performance of baseline substantially. In detail, by using our approach, we can achieve an improved performance of 3.43/4.65 ROPUGE-2 score, 2.33/3.18 BLEU score, and 3.68/4.28 METEOR score on Quora/Twitter dataset, outperforming the previous state-of-the-art model, which further demonstrates the effectiveness of our proposed approach.

\end{itemize}

Overall, combining the results of abstractive summarization, image captioning, video captioning, medical report generation, and paraphrase generation, the proposed multi-view decoding can well-perform the natural language generation, regardless of the downstream application scenarios.
Especially, our approach can outperform previous state-of-the-art models and set new state-of-the-art performances on image captioning, video captioning, medical report generation, and paraphrase generation, which further prove the effectiveness of our approach.

\begin{figure*}[t]

\centering
\includegraphics[width=0.65\linewidth,trim={0 -10 440 0},clip]{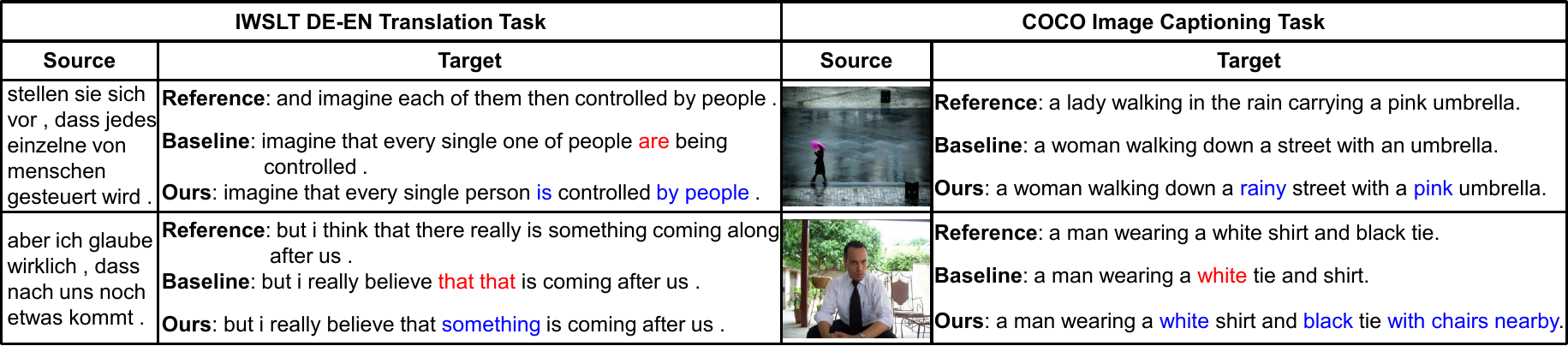} 
\includegraphics[width=0.65\linewidth,trim={437 0 0 0},clip]{example.pdf}
\caption{Examples of the target sentences generated by different methods. The color Blue denotes the examples when our model generates better target sentences than the baseline, while Red denotes unfavorable results.}
\label{fig:example}
\end{figure*}

\subsection{Case Study}
In Figure~\ref{fig:example}, we list some examples from DE-EN and COCO to analyze how our approach improves the baseline.
The examples show that our approach does not alter the structure of the output sentence significantly compared to the baselines. The reason is that our approach can be seen as an extension for fine-tuning existing models. However, our approach can capture more detailed information about the source sequence. 
For example, in machine translation, our approach enables the model to use words that are more precise, including verb forms, and singulars/plurals, especially when the baseline is unable to choose a proper word to continue the sentence, e.g., repetition. For the image captioning, our approach helps the model to generate more detailed captions in colors (e.g., \textit{pink} umbrella) and attributes (e.g., \textit{rainy} street), for each object. 
As we can see, the case study further proves our arguments and the effectiveness of the proposed approach.

\section{Conclusions}  
\label{sec:conclusion}
In this work, we focus on enhancing the information transfer between the encoder and the decoder for sequence-to-sequence learning, by injecting diverse source representations into the generation process. 
Different from the single view approach in existing work, we propose the \textit{layer-wise multi-view decoding} approach to route source representations of different granularity to different decoder layers, together with a global view preventing the regression of last encoder layers due to the \textit{hierarchy bypassing} problem. The multi-view decoding also builds upon continued training to mine the expressive power of encoder layer stack in learned conventional single-view models.
Out of several source-target routing strategies, we find that the granularity consistent attention (GCA) strategy for context attention shows the best improvements with the proposed layer-wise multi-view decoding.
Extensive experiments on diverse sequence generation tasks verify the effectiveness of our approach. 
In particular, it achieves new state-of-the-art results with almost negligible parameter increase on ten benchmark datasets, including two large-scale translation datasets, i.e., WMT EN-DE and WMT EN-FR, a low-resource IWSLT DE-EN translation dataset, a COCO image captioning dataset, two video captioning datasets, i.e., MSVD and MSR-VTT, two low-resource medical report generation datasets, i.e., MIMIC-CXR and IU-Xray, and two paraphrase generation datasets, i.e., Twitter and Quora.
The systematic analysis further proves the effectiveness and the generalization ability of our method, and particularly show that the use of different types of representations from the encoder, which provides different views of the source sequence, is necessary for improving the expressive power of the existing models.

\section*{Acknowledgments}
We thank all the anonymous reviewers and editors for their constructive comments and suggestions. Xu Sun is the corresponding author of this paper.

\bibliographystyle{ACM-Reference-Format}
\bibliography{sample-base}

\end{document}